\documentclass{article}
\usepackage{ijcai25}
\usepackage{times}
\usepackage{soul}
\usepackage{url}
\usepackage[hidelinks,colorlinks,
            linkcolor=blue, 
            anchorcolor=blue,
            citecolor=blue,]{hyperref}
\usepackage[utf8]{inputenc}
\usepackage[small]{caption}
\usepackage{graphicx}
\usepackage{amsmath}
\usepackage{amsthm}
\usepackage{booktabs}
\usepackage{algorithm}
\usepackage{algorithmic}
\usepackage{xspace}
\usepackage{color}
\usepackage{amssymb}
\usepackage[switch]{lineno}
\usepackage{subfigure}
\usepackage{subcaption}
\usepackage[table]{xcolor}
\usepackage{multirow}
\usepackage{enumitem}

\newcommand{\name}{\textsc{HiLo}\xspace}

\title{Rank Also Matters: Hierarchical Configuration for Mixture of Adapter Experts in LLM Fine-Tuning}

\author{
Peizhuang Cong
\and
Wenpu Liu
\and
Wenhan Yu
\and
Haochen Zhao
\And
Tong Yang
\\
\affiliations
Peking University\\
\emails
\{congpeizhuang, yangtong\}@pku.edu.cn
}
\begin{document}
\maketitle
\begin{abstract}
Large language models (LLMs) have demonstrated remarkable success across various tasks, accompanied by a continuous increase in their parameter size. Parameter-efficient fine-tuning (PEFT) methods, such as Low-Rank Adaptation (LoRA), address the challenges of fine-tuning LLMs by significantly reducing the number of trainable parameters. Recent studies have integrated LoRA with Mixture of Experts (MoE) architectures, leveraging multiple adapter experts and gating mechanisms to further improve fine-tuning performance. However, existing approaches primarily focus on adjusting the number of adapter experts per layer to optimize the size of introduced trainable parameters, while neglecting a critical factor of adapters' rank. 

To this end, we propose a hierarchical configuration scheme for adapter experts, \name, which flexibly adjusts the number and rank values of adapter experts across layers, matching the varying representational complexity cross model layers. Extensive experiments on multiple benchmark tasks demonstrate that \name outperforms existing methods in accuracy while introducing fewer trainable parameters, providing an efficient and practical solution for fine-tuning LLMs. 
\end{abstract}

\section{Introduction}\label{sec:introduction}
In recent years, the rapid advancement of the Large Langue Model (LLM) has demonstrated impressive performance on various tasks, e.g., natural language processing, computer vision, and multimodal applications \cite{liu2024gpt,chang2024survey}. 
However, as the parameter size of large models continues to scale up, substantial computational resources are required for model fine-tuning. In this context, parameter-efficient fine-tuning (PEFT) methods have emerged \cite{han2024parameter}, significantly reducing the number of trainable parameters while maintaining performance. 
PEFT techniques include adapter tuning, prompt tuning, prefix tuning, and Low-Rank Adaptation (LoRA) \cite{houlsby2019parameter,lester2021power,li2021prefix,hulora}. Among these, LoRA has gained considerable attention for its innovation of injecting low-rank matrices (referred to as adapters) into the model’s weight matrices, enabling efficient fine-tuning for downstream tasks without introducing trainable parameters size substantially. 

Based on LoRA, several studies have further optimized it by incorporating the architecture of Mixture of Experts (MoE) \cite{liu2024moe,zeng2024adamoe,dou2024loramoe}. MoE achieves superior computational efficiency and model capacity by employing multiple feed-forward networks (referred to as experts) to decompose complex problems into smaller subproblems, which will be processed by a subset of experts selected by a gating network \cite{zhou2022mixture}. 
The integration of LoRA and MoE termed the mixture of adapter experts, which involves introducing multiple adapters for model's weight matrices and activating a subset of them selectively via a gating network, resulting in an efficient and scalable fine-tuning solution. 

\begin{figure}[t]
    \centering
    \includegraphics[width=0.95\linewidth]{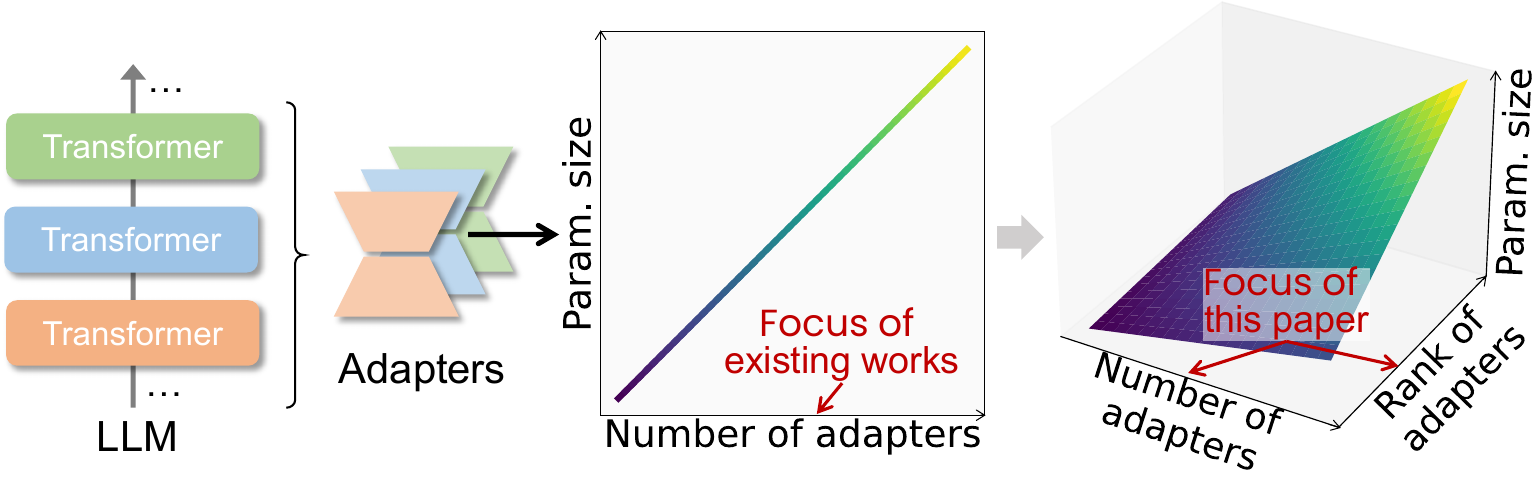}
    \caption{\textbf{Optimization focus of adapter experts}. Existing works primarily optimize the number of adapters, whereas this paper explores both the number and rank of adapters to improve fine-tuning performance.}\label{fig:param}
\end{figure}

The vanilla approach to constructing a mixture of adapter experts model involves assigning a fixed number of adapters to specific weight matrices of each layer, which neglects the variability and complexity across the layers of the pre-trained model. 
Then, some studies try to allocate a more suitable number of experts based on characteristic per model layer, aiming to further reduce trainable parameters while minimizing accuracy loss. 
For example, \cite{gao2024higher} assigns fewer experts to shallow layers and more to deeper layers, \cite{qing2024alphalora} adjusts the number of experts based on the training quality of model networks. 
However, all existing studies set a consistent rank for all adapters, which limits the flexibility and adaptability of adapter experts and results in unnecessary parameters across layers. 
In other words, as shown in Figure \ref{fig:param}, beyond the number of experts focused in existing studies, the rank configuration is also a critical factor in determining the size of introduced trainable parameters and influencing the feature learning capacity per adapter. 


To this end, we present \name, a hierarchical scheme for adapter experts configuration scheme, which aims to enhance the efficiency of mixture of adapter experts. Unlike existing studies that focus solely on the number of experts, \name holistically adjusts layer configurations by jointly considering the effects of expert numbers and ranks, aligning with the varying representational complexity of model layers in two dimensions. Extensive experimental results demonstrate that \name surpasses existing methods in accuracy while simultaneously achieving reductions in trainable parameters for fine-tune training and active parameters for inference. 

The main contributions of this paper are as follows:
\begin{itemize}
	\item We reveal that existing mixture-of-adapter-experts architectures focus solely on the number of adapter experts and experimentally demonstrate that the rank of adapters is also a critical factor for improving performance. 
    \item We propose a hierarchical configuration scheme, \name, which allocates different numbers and ranks to adapter experts across layers, enhancing the efficiency and performance of fine-tuning. 
    \item We conduct extensive experiments to validate that \name achieves higher model accuracy while reducing trainable and active parameters compared to existing methods.
\end{itemize}

\section{Related Work}\label{sec:relatedwork}
\subsection{Parameter Efficient Fine-Tuning}
The increasing size of LLM parameters motivates the development of parameter efficient fine-tuning \cite{han2024parameter}. Unlike full-parameter fine-tuning, PEFT methods selectively adjust or introduce a small number of trainable parameters without modifying the entire parameter set. Adapter tuning \cite{houlsby2019parameter}, prompt tuning \cite{lester2021power}, prefix tuning \cite{li2021prefix}, and LoRA \cite{hulora} are some representative PEFT schemes, and LoRA is one of the most widely employed strategies. It integrates the low-rank decomposition matrices into the weight update of the model and greatly reduces the number of trainable parameters. Specifically, the modified parameter matrix $w'\in \mathbb{R}^{n\times m}$ is computed as $w'=w+AB$, where $w \in \mathbb{R}^{n\times m}$ is the original parameter matrix, $A\in \mathbb{R}^{n\times r}$ and $B \in \mathbb{R}^{r\times m}$ are the low-rank decomposition matrices, and $r$ is much smaller than $n$, i.e., $r\ll min(n,m)$. 

Based on the basic LoRA, several optimizations have been proposed to align with the specific characteristics of LLMs and implementation requirements. The authors in \cite{dettmers2024qlora} further minimize parameter overhead by quantizing adapter parameters. LoRA+ \cite{hayoulora} applies differentiated learning rates based on the initial values of the adapter matrices $A$ and $B$, thereby improving fine-tuning performance in certain scenarios. Unlike conventional approaches, which typically initialize $A$ with random values sampled from a normal distribution and $B$ with zeros, VeRA \cite{kopiczkovera} initializes all of them with a normal distributed random values but freezes them, and adds a trainable vector for each of $A$ and $B$. It can further reduce the trainable parameters involved in fine-tuning with slight accuracy sacrifice.

\subsection{Mixture of Experts}
The mixture of experts architecture was first proposed by \cite{jacobs1991adaptive}, introducing an assignment mechanism based on input data to distribute tasks across multiple expert modules, thereby achieving efficient task specialization and model capacity utilization. 
With the growing popularity of the Transformer \cite{vaswani2017attention}, many studies revisited MoE by applying it to the corresponding Feed-Forward Network (FFN) layers, extending these layers into multiple expert networks. However, a prominent feature of the widely adopted MoE in transformer-based LLMs is the use of a sparse gating mechanism, which selects only a subset of experts for token processing, enabling LLMs to scale to an extreme scale. 

GShard \cite{lepikhin2020gshard} and Switch Transformer \cite{fedus2022switch} are pioneers that employ learnable top-2 or top-1 expert selection strategies, and DeepSeek \cite{dai2024deepseekmoe} implements a shared expert isolation scheme. HashLayer \cite{roller2021hash} uses a hashing-based way to select experts for tokens, improving the stability of model training. \cite{huang2024harder,zhou2022mixture,yang2024xmoe} allow different numbers of experts to be assigned for different tokens, enhancing model flexibility. Besides studies on architectures and training strategies of MoE, recent years have also witnessed the emergence of many MoE-based multimodal models \cite{riquelme2021scaling,mustafa2022multimodal,du2022glam}.

\subsection{LoRA Meets MoE}
Based on basic LoRA technology, some studies have further introduced the MoE architecture, where each weight matrix's LoRA adapter is no longer a single module but are a set of adapters controlled by a gating network \cite{yang2024moral,wu2024parameter}. Such the approach enhances the capability of the fine-tuned model while maintaining scalability \cite{li2024mixlora,wumixture}. 

MoELoRA \cite{liu2024moe} combines the strengths of these two techniques to achieve an efficient multi-task fine-tuning framework. 
AdaMoE \cite{zeng2024adamoe} ingeniously introduces additional null expert networks, enabling a learnable and dynamic selection of the experts. 
LoRAMoE \cite{dou2024loramoe} classifies all experts of each layer into two categories, assigning them to process either world knowledge or fine-tuning knowledge, thereby mitigating the issue of world knowledge forgetting of the fine-tuning. 
MoLA \cite{gao2024higher} implements a layer-wise expert allocation strategy and demonstrates through experimental results that deeper networks require more experts than shallower ones.

\section{Motivation}\label{sec:motivation}
\subsection{Observation and Analysis}\label{subsec:observation}
\begin{figure}[t]
  \centering
  \subfigure[Distribution of output values]{
    \label{fig:dis}
    \includegraphics[width=0.97\linewidth]{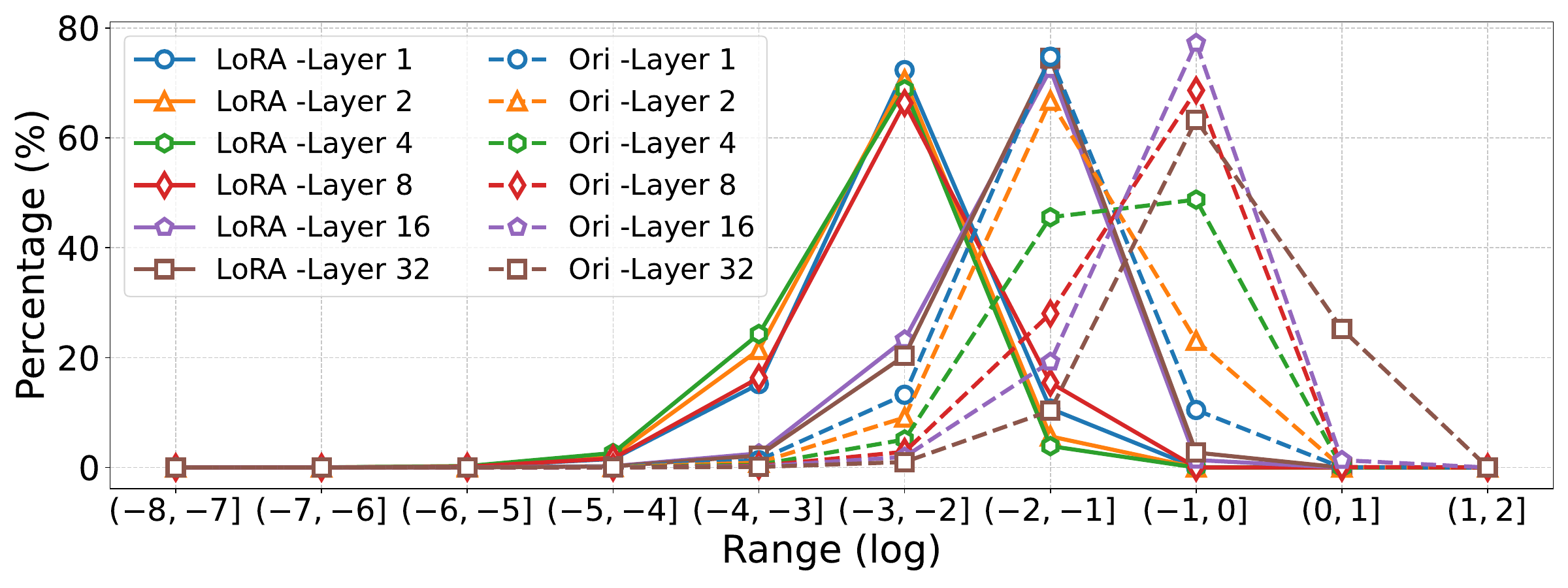}
  }
  \subfigure[Cumulative distribution of output values]{
    \label{fig:cumdis}
    \includegraphics[width=0.97\linewidth]{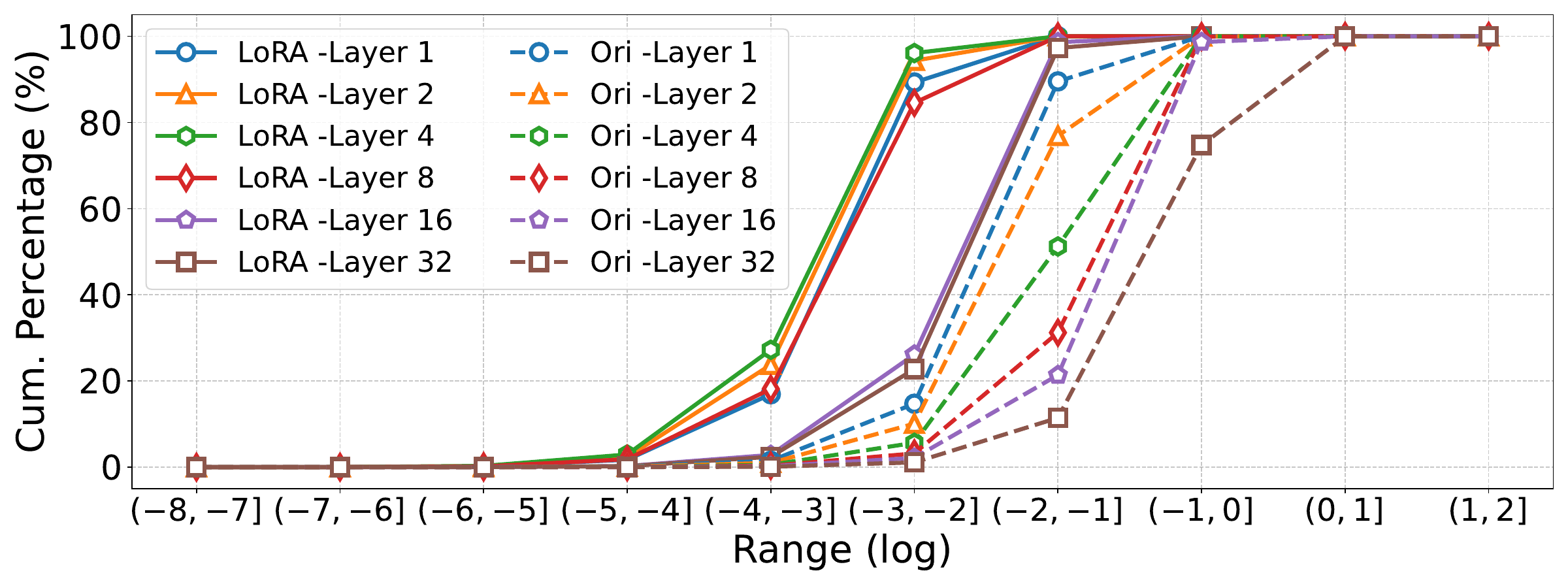}
  }
  \caption{Analysis of output values between original model weight matrices and adapters cross layers}\label{fig:distribution}
\end{figure}

Without loss of generality, in the case of a particular layer of the model's network, LoRA-based fine-tuning involves adding the two output vectors of the original network and the LoRA adapter, the resulting vector will be the final output of this layer and subsequently fed into the next layer for further computation, i.e., $y=wx+\Delta wx$, where $w$ is the original weight matrix and $\Delta w$ is the adapter. Similarly, in the mixture of adapter experts architecture, the output vectors of multiple adapter experts are performed weighted summation according to the activation states, which will be added to the output vector of the original network. 

The outputs of original networks and adapter experts are tracked during model inference for further investigation. The model configuration is as followings: using Llama\,2-7B \cite{touvron2023llama} as the base model, allocating each network layer with 8 adapter experts of all ranks 8, employing the Top-2 activation strategy, and fine-tuned on the ScienceQA dataset \cite{lu2022learn}. When inputting a randomly selected subset of the test samples into the model, the distributions of the output values of Feed-Forward Neural Network (FFN) layers and corresponding adapter expert layers are presented in Figure \ref{fig:distribution}. 

\begin{figure}[t]
    \centering
    \includegraphics[width=0.95\linewidth]{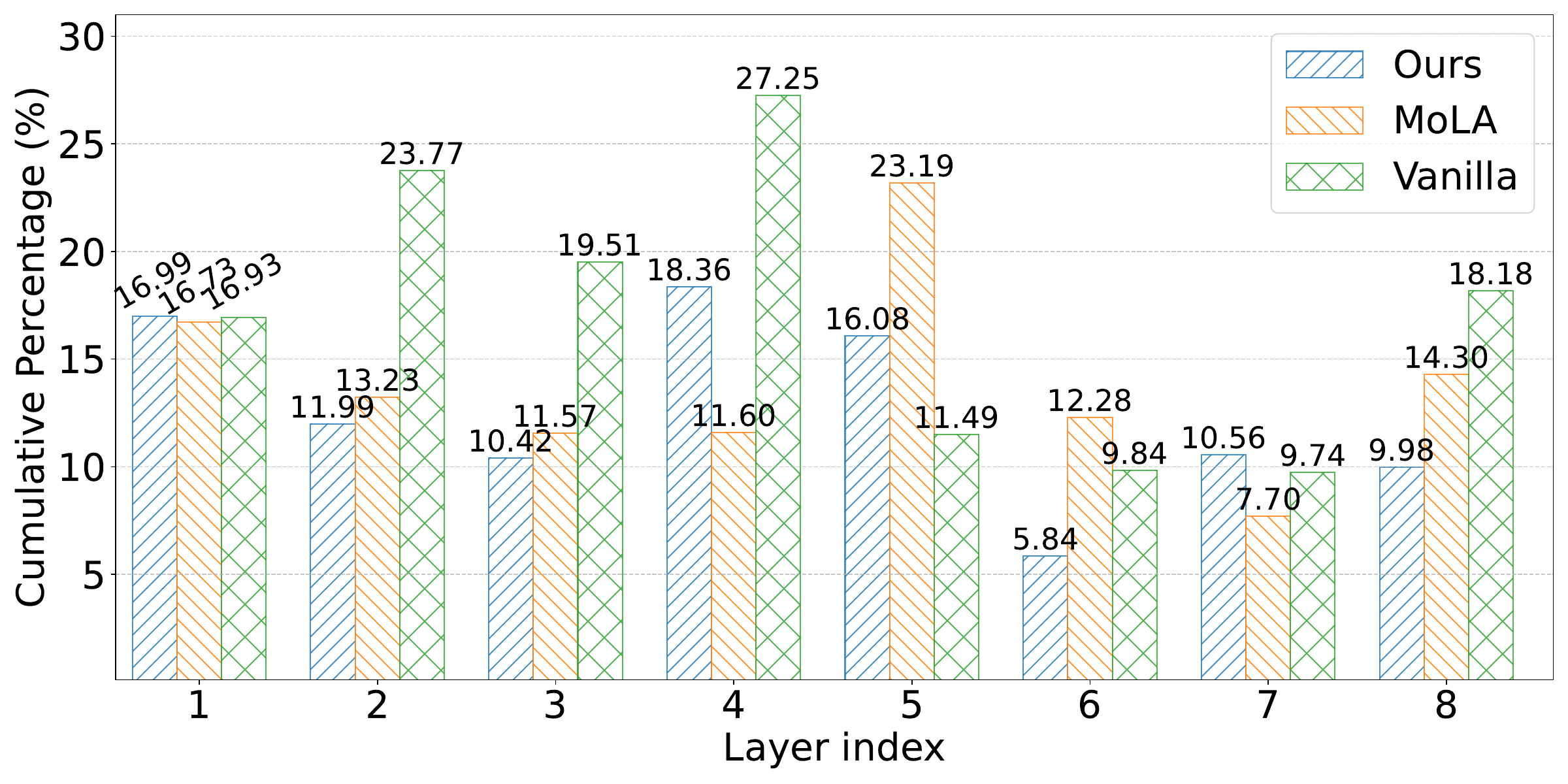}
    \caption{\textbf{Proportions of values (\textless$10^{-3}$) of vanilla method, MoLA, and \name}. (1) the proportions of three methods in the 1st layer are very close; (2) except 1st layer, MoLA has lower proportions than Vanilla and \name in 4 and 2 layers, respectively; (3) \name exhibits lower proportions than Vanilla and MoLA in 5 layers.}
    \label{fig:3bar}
\end{figure}

The following trends be observed: 1) the output of the original network exceeds the output of its adapter experts by more than one or two orders of magnitude in the range with the highest concentration of values within the same network layer; 2) for the adapter experts’ outputs, shallow layers exhibit a higher proportion of small values close to zero compared to deep layers. These trends occur because the shallow layers of the model typically perform general feature extraction, while the deeper layers are responsible for learning specialized features. For a new fine-tuning task, the original networks in the shallow layers are capable of performing basic feature extraction to a certain extent, thereby minimizing the need for fine-tuning these layers. Therefore, with the same size of trainable parameters, allocating fewer adapter experts to the shallow layer and more adapter experts to the deep layer, rather than equally, is a straightforward and effective allocation strategy to maximize the effectiveness of such parameters. The output values distribution of MoLA (i.e., assigning fewer adapters to shallow layers) under the same fine-tuning settings, except for the adapter allocations, are present in Figure \ref{fig:3bar}. Compared to the equal allocation of experts, such optimized allocation way reduces the proportion of small output values from shallow-layer experts. 

However, in MoE architectures, assuming all other configurations remain unchanged, the number of experts within a reasonable range is generally expected to exhibit a positive correlation with the capability of this MoE layer. From this perspective, strategies that directly reduce the number of adapter experts fail to consider the influence of the individual expert’s capacity. On the one hand, the size of trainable parameters is not only proportional to the number of experts but also to the rank of the adapter experts. The rank determines the parameter size of each adapter expert, which directly impacts its fitting ability. On the other hand, when the number of experts is fixed, there is an upper limit to improving the fitting ability of this MoE layer by simply increasing the parameter size of experts. Specifically, once an individual expert’s fitting ability exceeds the layer’s requirements, further increasing its parameter size provides diminishing returns about model capability. In summary, the number of experts and the capacity of individual experts jointly influence the overall performance of the MoE layer.

\begin{figure*}
    \centering
    \includegraphics[width=0.8\linewidth]{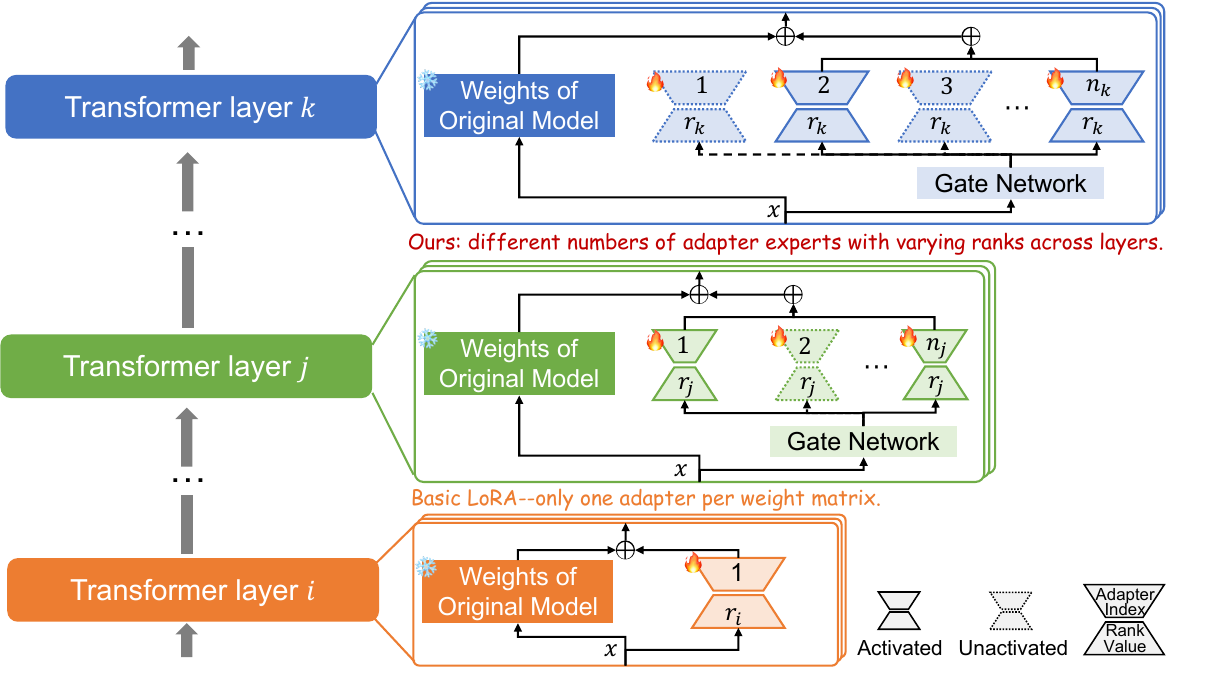}
    \caption{\name Design}\label{fig:framework}
    \vspace{-5pt}
\end{figure*}

\subsection{Key Idea}\label{subsec:motivation}
Given the characteristic of the shallower to deeper layers of the model, there are requirement differences cross layers in the capability of individual adapter experts for fine-tuning. It exhibits incremental requirements for adapter experts from shallow to deep layers in fitting capability. And the parameters size, which decided by the rank of adapter, scales to the fitting capability. 

Therefore, inspired by the analysis above, this paper proposes a hierarchical rank setting scheme to adapter experts for efficient LLM fine-tuning, \name. When maintaining the same trainable parameter size with MoLA, the proportions of output values smaller than $10^{-3}$ in the shallow layers (layers 1 to 8) of \name are presented in Figure \ref{fig:3bar}. It indicates that \name can further reduce the proportion of such small values than MoLA. 

It notes that the final goal of model fine-tuning is not to minimize such proportions but to improve model accuracy or reduce trainable parameter size. We aim to provide an analytical perspective by this data analysis. The details and outperformances of \name will be presented in Section \ref{sec:methodology} and \ref{sec:evalution}, respectively.

\section{\name Design}\label{sec:methodology}
In this section, we first describe the extension of the basic LoRA method to the MoE architecture, and then present the adapter experts configuration about allocation and activation of experts. Finally, we introduce a simple and efficient rank-setting strategy, which can further reducing parameters while improving accuracy of the fine-tuned model.

\subsection{Mixture of Adapter Experts}
Extending LoRA to MoE architectures is to transform the original single adapter, which comprises one pair of low rank matrices \(A\) and \(B\), into multiple adapter experts. 
Suppose there is a model with \(N\) Transformer layers and each layer \(i\) has \(E_i\) adapter experts, denoted by \(e^j_i\), where \(j \in E_i\). 
In line with the basic LoRA initialization scheme, each adapter expert has its matrix \(A\) initialized with a random Gaussian distribution and \(B\) with zeros.
Moreover, a gate network \(G_i\) is introduced for layer \(i\). It is responsible for analyzing the features of each input token \(x\) and computing activation probabilities \(P^j_i\) for all adapter expert \(e^j_i\) via a softmax function. 

Following the MoE computation procedure with a Top-K activation policy, for a given token \(x\), the adapter experts \(e^i_k, k\in K\) with highest \(K\) probabilities are activated. Then, the output of this mixture of adapter experts layer can be expressed as:
\[
{Output}_{e_i} 
= \sum_{k \in K} p^k_i \, w^k_i \, {x},
\quad
p^k_i 
= \frac{P^k_i}{\sum_{j \in K} P^j_i},
\]
where \(w^k_i\) represents the parameter matrix of the expert \(e^k_i\). 
Assuming \(W_i\) is the original weight matrix of layer \(i\), the final output of this layer yields:
\[{Output}_i = W_i \, {x} + {Output}_{e_i}.\] 

So far, it is possible to select a subset of adapter experts dynamically, enhancing both the capacity and adaptation flexibility of LoRA. Meanwhile, \({Output}_{e_i}\) is fed into the next layer as the input.

\subsection{Adapter Experts Setting}
In this subsection, we present the experts allocation strategies and activation policies, which determine sizes of trainable and active parameters, respectively, thereby affecting the model’s efficiency.

\subsubsection{Dynamic Experts Allocation}
The vanilla allocation strategy assigns the same number of adapter experts to each layer. However, as mentioned earlier, such an approach overlooks the differences across layers, resulting in inefficiencies such as introducing unnecessary trainable parameters or failing to utilize them fully. Two proven flexible expert allocation strategies are presented here \cite{gao2024higher,qing2024alphalora}. First, based on the varying complexity of processing tasks across different layers, experts can be allocated incrementally from shallow to deep layers. This intuitive method demonstrates performance improvements compared to a fixed expert allocation. Second, the Heavy-Tailed Self-Regularization theory can be used to evaluate the training quality of the original model layers, guiding a flexible and adaptive allocation of experts to each layer.

\subsubsection{Activation of Experts}
For expert activation, Top-$K$ is a feasible and simple manner. However, activating the fixed number of experts neglects the processing complexity among different tokens, which may result in either redundancy or insufficiency in processing for certain tokens. 

Comparatively, learning-based policies enable the activation of a variable number of experts. 
The Top-$P$ series methods exemplify this policy by activating experts for each token, either by selecting in order of decreasing probability until the cumulative probability exceeds a predefined threshold $P$, or by directly selecting all experts whose probabilities surpass the threshold. $P$ is predefined, and the loss function incorporates constraints on the number of activated experts. Additionally, another implementation involves inserting several placeholder adapters without parameters into the adapter expert layer \cite{zeng2024adamoe}. These placeholders do not perform computations on the token but allow the model to activate a varying number of true experts for each token while still adhering to the Top-$K$ policy. For instance, if $n$ of the selected experts are placeholder adapters, then, only $K$-$n$ true experts are actually activated. During training, the model learns to activate the appropriate number of experts dynamically for each token.

These above two kinds of mainstream policies for expert activation will be evaluated in comparative experiments.

\subsection{Hierarchical Rank-setting}
In the mixture of adapters experts architecture, the number of adapter experts per layer and their rank are critical factors influencing model performance and trainable parameter size. The rank of an adapter expert directly determines its representational ability and parameter size, while the current layer-wise expert's allocation strategy leaves room for further optimization. That is, the rank of shallow-layer adapter experts can be set lower than that of deeper layers, which further satisfies diverse representational requirements and improves parameter utilization. 

Formally, with the layer as the rank setting granularity, the number of adapter experts in each layer can be expressed as: 
\[r_i = r_{min} + \left\lfloor \frac{r_{max} - r_{min}}{\left\lceil \frac{n}{l} \right\rceil} \right\rfloor \cdot \left\lfloor \frac{i-1}{l} \right\rfloor,\]
where \(\left\lceil \cdot \right\rceil\) and \(\left\lfloor \cdot \right\rfloor\) respectively represent the rounding up and rounding down operations; \(r_{min}\) and \(r_{max}\) respectively denote the minimum and maximum rank values. Typically, \(r_{min}\) is set to 2, depending on the model’s scale, and \(r_{max}\) can take values such as 8, 16, 32, or higher. \(n\) indicates the total layers of the fine-tuned model. Optionally, it is feasible to set same rank for for every \(l\) layers. 

Moreover, considering the computational characteristics of hardware devices, rank values can be flexibly configured as multiples or powers of 2 based on the formula in practical implementation, rather than strictly adhering to it.
\begin{table*}[t]
\centering
\resizebox{\textwidth}{!}{
    \begin{tabular}{lcccccccccccc}
    \toprule
    & \multicolumn{3}{c}{Adapter Expert Setting} & \multicolumn{2}{c}{Parameter $\downarrow$} & \multicolumn{6}{c}{Accuracy (\%) $\uparrow$} \\
    \cmidrule(lr){2-4} \cmidrule(lr){5-6} \cmidrule(lr){7-13}
    \multirow{-2}{*}{Method}& {Exp.} & {Rank} & {Top-$K$} & {Active} & {Trainable} & {SQA} & {CQA} & {OQA} & {MRPC} & {COLA}  & {RTE} & {Avg.}\\
    \midrule
    Llama\,2 & - & - & - & - & - & 40.23 & 20.58 & 25.20 & 66.55 & 55.99  & 53.42 & 43.66\\
    \midrule
    Vanilla  & 8  & 8  & 2 & 2 & 1 & 90.10 & 78.21 & 76.20 & 85.15 & 84.85  & 87.72 & 83.71\\
    \midrule
    MoLA & $\triangledown$ & 8 & 2 & 2 & 0.63 & 89.61 & \textbf{76.00} & 74.20 & 84.23 & 85.52  & 84.47 & 82.34\\
    \rowcolor{gray!15}{\name} & 8  & $\triangledown$ & 2 & \textbf{1.25} & 0.63 & \textbf{90.42} & 75.92 & \textbf{76.20} & \textbf{84.52} & \textbf{86.19} & \textbf{84.48} & \textbf{82.96}\\
    \midrule
    AlphLoRA & $ \ddagger$ & 8 & 2 & 2 & 0.63 & 89.70 & 75.83 & 74.60 & 82.49 & 85.13 & 85.55 & 82.22\\
    \rowcolor{gray!15}\name & $\ddagger\ddagger$ & $\triangledown$ & 2 & \textbf{1.25} & 0.63 & \textbf{90.15} & \textbf{76.58} & \textbf{75.80} & \textbf{84.46} & \textbf{84.27} & \textbf{88.09} & \textbf{83.23}\\
    \bottomrule
    \end{tabular}
    }
    \caption{Performances between \fcolorbox{white}{gray!15}{\name (ours)} and comparative methods}\label{tab:method_comparison1}
    \vspace{-3pt}
\end{table*}

\section{Evaluation}\label{sec:evalution}
In this section, we demonstrate the outperformance of \name over various benchmarks, mainly in reducing trainable or active parameter size and enhancing fined-tuned model performance through its hierarchical rank setting.

\subsection{Experiment Setup}
\textbf{Model and Dataset}. 
The Llama\,2-7B \cite{touvron2023llama} is selected as the base model of fine-tuning due to its high deployment and popularity within the LLM community. To evaluate the performances of \name and comparison methods, two distinct types of tasks are employed using widely recognized datasets. The first type of task involves commonsense reasoning and includes the following dataset: ScienceQA (SQA) \cite{lu2022learn}, CommonsenseQA (CQA) \cite{talmor2019commonsenseqa}, and OpenBookQA (OQA) \cite{mihaylov2018can}. The second type of task concentrates on semantic understanding, which uses three datasets from the renowned GLUE Benchmark \cite{wang2019glue}: the Microsoft Research Paraphrase Corpus (MRPC), the Recognizing Textual Entailment (RTE), and the Corpus of Linguistic Acceptability (COLA).

\textbf{Baselines}. 
The pre-trained Llama\,2-7B model and vanilla framework of the mixture of adapter experts are selected as comparison methods. Specially, the Llama\,2-7B model is evaluated using prompt engineering tailored for each dataset, and the vanilla method allocates 8 experts with a rank of 8 for each weight matrix of every Transformer layer. 

Based on the vanilla method, the optimization schemes of adapter experts of existing studies can be divided into three categories: 
\begin{enumerate}[label=(\alph*)]
\item The first one is the manual-based expert allocation represented by MoLA \cite{gao2024higher}. This study indicates that in Llama\,2-7B-based fine-tuning, introducing the same size of trainable parameters, allocating 2, 4, 6, and 8 experts in groups of every 8 layers from shallow to deep layers is the performance-optimal allocation. To facilitate the description, this allocation strategy is denoted as “$\triangledown$” in this paper, which serves as a comparison method. 

\item The second one is AlphaLoRA \cite{qing2024alphalora}, which allocates experts based on the training quality of the original model. When introducing the same size of trainable parameters as MoLA, it has a layer-wise allocation strategy\footnote{For the LLaMA\,2-7B model, AlphaLoRA allocates the number of experts to each Transformer layer as follows: 1, 3, 4, 4, 4, 4, 4, 3, 3, 3, 3, 3, 2, 2, 3, 3, 3, 3, 3, 4, 4, 5, 5, 7, 6, 8, 9, 8, 7, 8, 6, 6, 8, 7, 7, 9, and 5, respectively.} denoted as “$\ddagger$” in this paper, which is another comparison method. Both of these approaches strive to improve the performance of the fine-tuned model with the limited introduced trainable parameters. 

\item Additionally, another category of methods improves model efficiency by flexibly adjusting the number of activated experts to reduce the size of activated parameters. For example, AdaMoE \cite{zeng2024adamoe} achieves this by introducing placeholder experts that neither introduce trainable parameters nor incur activation overhead. In the experiments conducted in this paper, a 1:1 ratio of real experts to placeholder experts is set for each layer, which is also employed as a baseline for comparison. 
\end{enumerate}

\name is integrated into these aforementioned comparison methods. Except for the rank, which follows \name, all other settings remain unchanged with corresponding comparison method. The effectiveness of the proposed scheme can be validated by performance improvements in terms of trainable parameters, active parameters, and accuracy.

\subsection{Experiment Results}
The trainable parameters introduced by vanilla, which mainly involves all adapters with the rank of 8, is defined as a unit of 1. The active parameters that vanilla method activates 2 adapters with rank of 8 for each token is defined as 2. Other methods are scaled proportionally based on the number of introduced or activated adapters and corresponding rank values. All experiments performed on 4 NVIDIA 4090 GPUs.

\subsubsection{Performance Evaluation}
Except for the pre-trained Llama\,2 and Vanilla method, the trainable and active parameters of MoLA and AlphaLoRA are constant, and the active parameters corresponding to AdaMoE are dynamic. To this end, the comparison experiments are conducted respectively, and the specific comparison results are as follows. 

\textbf{(1) \name vs. MoLA \& AlphaLoRA}. MoLA reduces the total number of adapters in the model while keeping the adapter rank unchanged, thereby reducing the trainable parameters to 0.63 compared to Vanilla. To ensure fairness in the comparison, the trainable parameters introduced by AlphaLoRA are kept consistent with those of MoLA. Furthermore, since MoLA and AlphaLoRA also adopt the Top-2 activation scheme, their active parameters remain the same as Vanilla's.

Based on this, \name reduces the rank values progressively from deep to shallow layers, grouped every eight layers, to 8, 4, 6, and 2. To ensure fairness of comparison, the total trainable parameter size of \name is kept consistent with MoLA and AlphaLoRA by proportionally increasing the number of adapter experts from shallow to deep layers. Compared to MoLA, \name assigns eight experts per layer, while compared to AlphaLoRA, the number of experts per layer in \name is calculated as Rank$_i$/8*AlphaLoRA$_i$, where Rank$_i$ represents the rank value of adapters in layer $i$, and AlphaLoRA$_i$ represents the number of adapter experts in layer $i$ of AlphaLoRA, denoted as `$\ddagger\ddagger$' in this paper. In terms of active parameters, although \name adopts Top-2 activation scheme, the decreasing rank result in proportionally fewer active parameters in the shallow layers. Consequently, the active parameters of \name are reduced from 2, as in Vanilla, MoLA, and AlphaLoRA, to 1.25, representing a 37.50\% reduction. 

In terms of accuracy, since Llama\,2 is not fine-tuned on new datasets, it demonstrates relatively low performance. The Vanilla achieves higher accuracy by increasing the size of trainable and active parameters as the cost. However, \name, with a smaller parameter size, still outperforms the Vanilla method on two datasets. Comparing MoLA and AlphaLoRA, as shown in Table \ref{tab:method_comparison1}, \name is lower (0.08\%$\downarrow$) than MoLA only on one of six datasets (CQA) while achieving a maximum improvement of 2.00\% on the OQA. Moreover, \name outperforms AlphaLoRA across all datasets, with the highest improvement of 1.97\% on the MRPC dataset and an average improvement of 1.01\%.

The experimental results demonstrate that \name achieves higher model accuracy compared to MoLA and AlphaLoRA while introducing the same size of trainable parameters and the lower size of active parameters.

\begin{table}[t]
    \centering
    \resizebox{0.9\linewidth}{!}{
    \begin{tabular}{lc>{\columncolor{gray!15}}cc>{\columncolor{gray!15}}c}
    \toprule
     & \multicolumn{2}{c}{Active Parameter $\downarrow$} & \multicolumn{2}{c}{Accuracy (\%) $\uparrow$}\\
    \cmidrule(lr){2-3} \cmidrule(lr){4-5}
    \multirow{-2}{*}{Dataset} & AdaMoE & \name & AdaMoE & \name\\
    \midrule
    {SQA}  & 1.05 & \textbf{0.68} & 89.43 & \textbf{90.06}\\
    {CQA}  & 1.01 & \textbf{0.63} & 75.26 & \textbf{75.43}\\
    {OQA}  & 1.03 & \textbf{0.63} & 72.60 & \textbf{73.60}\\
    {MRPC} & 1.03 & \textbf{0.64} & 84.81 & \textbf{85.62}\\
    {COLA} & 1.01 & \textbf{0.63} & 85.42 & \textbf{85.81}\\
    {RTE}  & 1.09 & \textbf{0.67} & 81.94 & \textbf{85.20}\\
    \midrule
    {Avg.} & 1.04 & \textbf{0.65} & 81.58 & \textbf{82.62}\\
    \bottomrule
    \end{tabular}
    }
    \caption{\textbf{Performances of AdaMoE and \name}. Allocating 8 experts and 8 placeholder experts to each layer and employ a Top-2 activation strategy. AdaMoE's adapter rank is 8 and \name sets the rank to $\triangledown$ across layers.}
    \label{tab:method_comparison2}
    \vspace{-10pt}
\end{table}

\textbf{(2) \name vs. AdaMoE}. 
To compare with AdaMoE, \name reduces the rank values from deep to shallow layers, grouped every 8 layers, to 8, 4, 6, and 2, while keeping the number of experts constant. In contrast, the activations of real and placeholder adapter experts are learned during model fine-tuning. Similarly, the trainable parameter size of AdaMoE is 1, whereas \name's is 0.63.

In addition, we tracked the expert activation during validation. As shown in Table \ref{tab:method_comparison1}, the average active parameter of AdaMoE across datasets is 1.04, while \name's is 0.65, representing a 37.5\% reduction. For accuracy, \name outperforms AdaMoE across all datasets, with the highest improvement of 3.26\% on RTE dataset and an average improvement of 1.04\%. 

The experimental results exhibit that \name achieves higher model accuracy with fewer both trainable and active parameters compared to AdaMoE.

\subsubsection{Ablation Studies}
To further evaluate the performance of \name, we conducted another comparison with MoLA across all datasets. In this experiment, it does not constrain the introduced trainable parameters to be the same size. Instead, \name configures both the number of adapter experts and values of adapters' ranks as $\triangledown$. With this setup, the total size of trainable parameters in \name is reduced to 0.39, indicating a 37.50\% decrease compared to MoLA. Similarly, the size of active parameters in \name is reduced to 1.25, reflecting a 37.50\% reduction from MoLA's 2. 

\begin{figure}[t]
    \centering
    \includegraphics[width=0.9\linewidth]{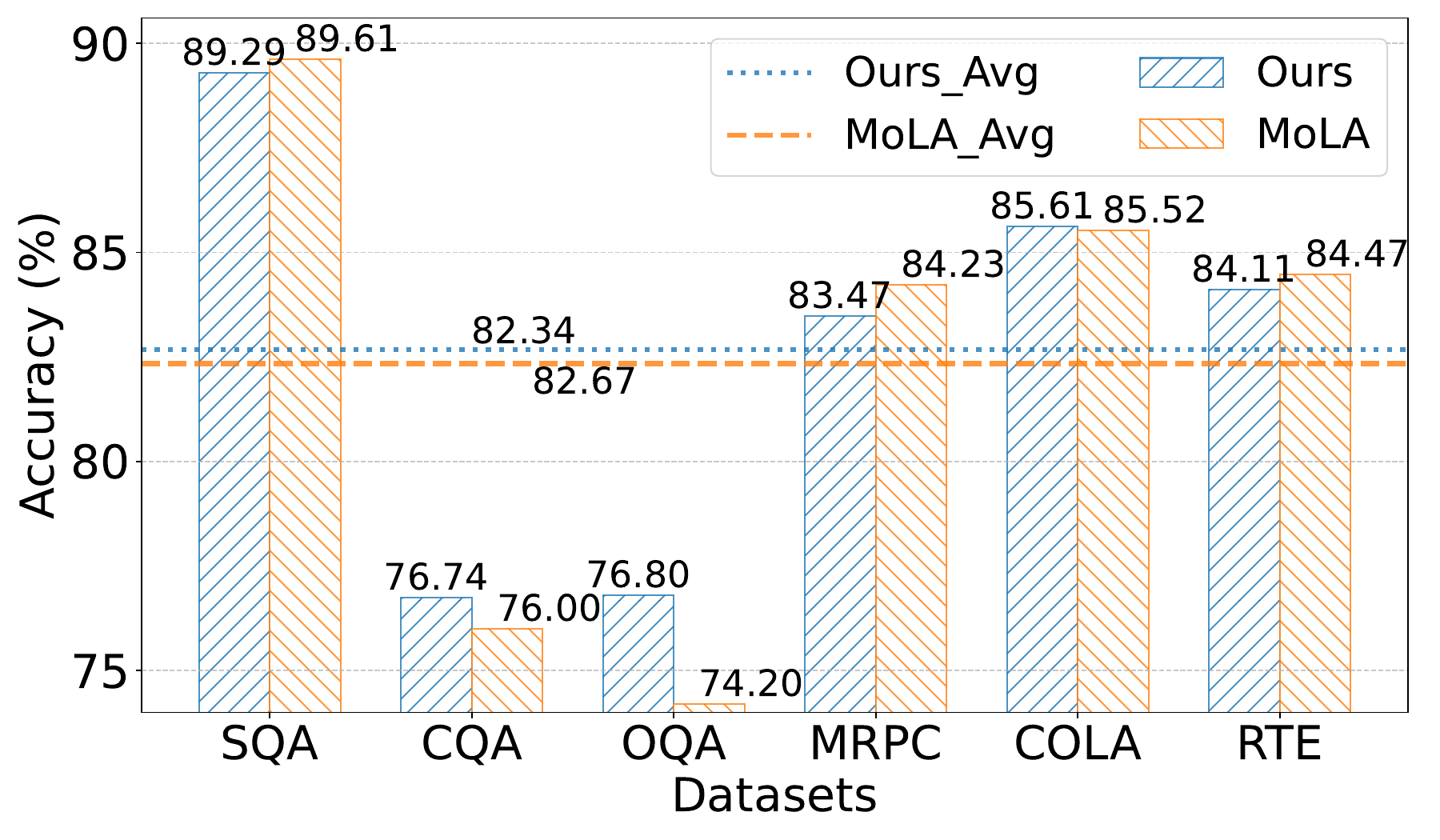}
    \caption{\textbf{Accuracy comparison between \name and MoLA}. \name assigns experts and rank as $\triangledown$ across layers.}
    \label{fig:comparsion_3}
    \vspace{-5pt}
\end{figure}

In terms of accuracy, as shown in Figure \ref{fig:comparsion_3}, \name outperforms MoLA on the CQA, OQA, and COLA datasets and achieves higher average accuracy across all datasets. These results further demonstrate that \name can achieve higher fine-tuning accuracy with fewer trainable and active parameters compared to MoLA. 

\begin{table}[t]
\centering
\resizebox{0.9\linewidth}{!}{
    \begin{tabular}{ccccc}
    \toprule
    \multicolumn{2}{c}{Adapter Experts} & \multicolumn{2}{c}{Parameter $\downarrow$} &  \\
    \cmidrule(lr){1-2} \cmidrule(lr){3-4}
    {Rank} & {Top-$K$} & {Active} & {Trainable} & \multirow{-2}{*}{{Accuracy (\%) $\uparrow$}} \\
    \midrule
    8 & 2 & 1.03 & 0.63 & 88.44 \\
    \rowcolor{gray!15}$\triangledown$ & 2 & \textbf{0.77} & \textbf{0.39} & \textbf{89.74} \\
    \midrule
    8 & $\triangledown$ & 1.65 & 0.63 & \textbf{90.56} \\
    \rowcolor{gray!15}$\triangledown$ & $\triangledown$ & \textbf{1.23} & \textbf{0.39} & 90.24 \\
    \bottomrule
    \end{tabular}
    }
    \caption{\textbf{Performances of Mix and \name in SQA}. Mix and \name allocate experts and placeholder expert as $\triangledown$ across layers.}
    \label{tab:ablation_experiments}
    \vspace{-5pt}
\end{table}

We integrate MoLA and AdaMoE as a new method,  denoted by Mix, which combines layer-wise expert allocation and learning-based adapter experts activation. On this basis, we introduce \name as a comparative method based on Mix, with experiments conducted on the SQA dataset. Under the condition that trainable parameters are unaffected by the activation scheme, the trainable parameters of Mix and \name are 0.63 and 0.39, respectively, with \name achieving a 37.50\% reduction. 

According to statistical analysis, when the adapter experts activation is the Top-2 scheme, the active parameters of Mix and \name are 1.03 and 0.77, respectively, representing a 25.24\% reduction. Although the Top-$\triangledown$ activation is not practically applied, we conducted validation and evaluation under this setting, where Mix and \name exhibit active parameters of 1.65 and 1.23, respectively, resulting in a 25.45\% reduction. 

In terms of accuracy, as shown in Table \ref{tab:ablation_experiments}, under the de facto widely employed Top-2 activation scheme, \name outperforms Mix with a 1.30\% improvement. Under the Top-$\triangledown$, \name’s accuracy is 0.32\% lower than Mix, but it still achieves higher accuracy than Vanilla, MoLA, and AlphaLoRA, while reducing both trainable and active parameters simultaneously.

\subsubsection{Exploratory Studies}

\begin{figure}
    \centering
    \includegraphics[width=1\linewidth]{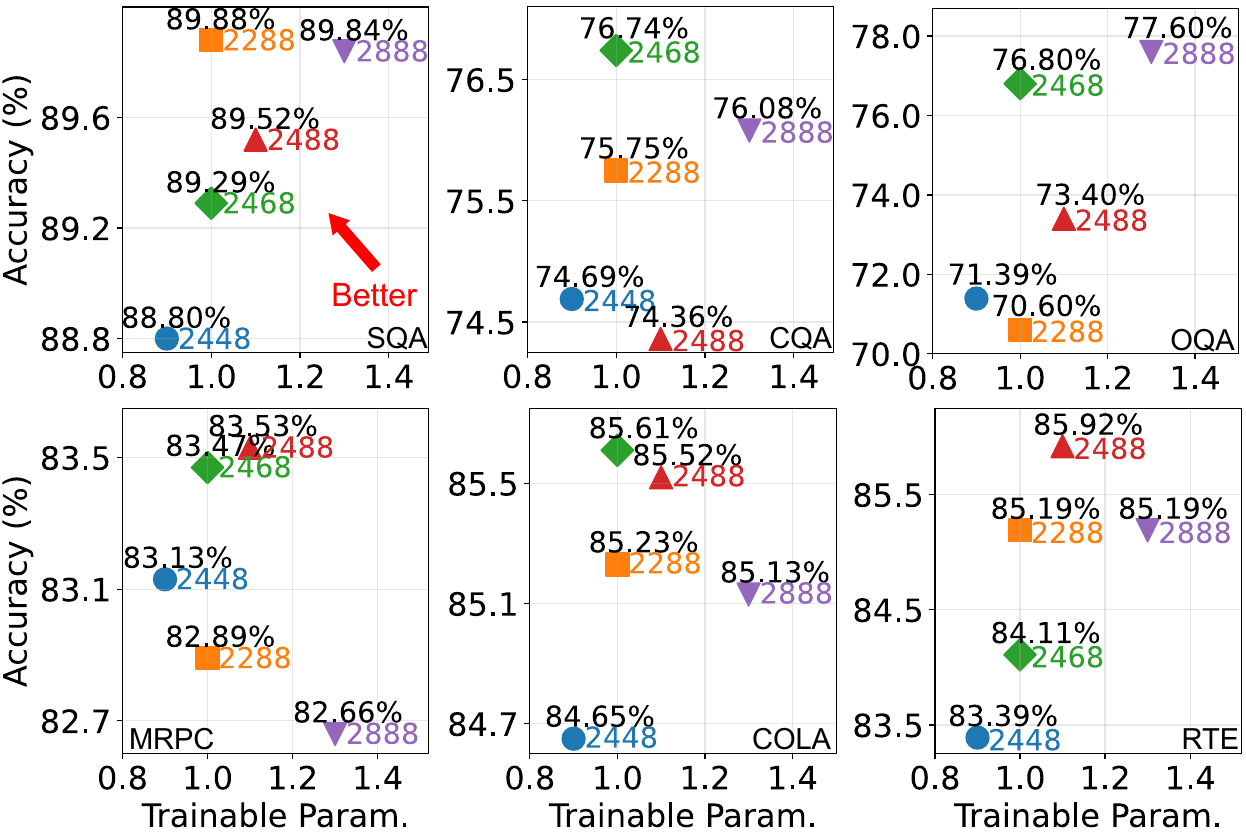}
    \caption{Accuracy performances of \name with different ranks.}\label{fig:rank_config}
    \vspace{-5pt}
\end{figure}

\name is evaluated by various rank configurations, specifically grouping every eight layers with ranks set as 2448, 2288, 2468, 2488, and 2888 from shallow to deep layers, with the corresponding trainable parameters gradually increasing. The parameter size of 2468 configuration is assigned unit 1. The accuracy results for these five configurations across six datasets are shown in Figure \ref{fig:rank_config}. It can be observed that the configuration of 2468 (green $\blacklozenge$) achieves the best balance between parameter size and accuracy. This configuration is also adopted in the above experiments.
\section{Conclusion}\label{sec:conclusion}
This paper proposes \name, a hierarchical configuration scheme for the mixture of adapter experts in LLM fine-tuning, enabling flexible adjustments to both the number and rank of adapter experts to better align with the varying representational complexity across model layers. Extensive experimental results demonstrate that \name outperforms existing methods in accuracy while achieving reductions in trainable and active parameters across diverse datasets.
\bibliographystyle{named}
\bibliography{ijcai}

\end{document}